\pdfoutput=1
\documentclass[lettersize,twoside,journal]{IEEEtran}
\usepackage{amsmath,amsfonts}
\usepackage{algorithmic}
\usepackage{algorithm}
\usepackage{array}
\usepackage[caption=false,font=normalsize,labelfont=sf,textfont=sf]{subfig}
\usepackage{textcomp}
\usepackage{stfloats}
\usepackage{url}
\usepackage{verbatim}
\usepackage{graphicx}
\usepackage{cite}
\usepackage{mathrsfs}
\usepackage{color}
\usepackage{colortbl}
\definecolor{Ocean}{RGB}{129,154,254}
\definecolor{orange}{RGB}{254,144,95}
\usepackage{bbding}
\usepackage{pifont}
\usepackage{tabularx}
\usepackage{multirow}
\usepackage{diagbox}
\usepackage{xcolor}
\usepackage[utf8]{inputenc}  
\usepackage[T1]{fontenc}     

\usepackage{multirow}
\usepackage{booktabs}
\usepackage{graphicx}
\usepackage{adjustbox}

\begin{document}
\title{Semi-Supervised 360 Layout Estimation with Panoramic Collaborative Perturbations}

\author{ Junsong~Zhang, Chunyu~Lin,
Zhijie~Shen, Lang Nie,
Kang~Liao, Yao~Zhao,~\IEEEmembership{Fellow,~IEEE}
\thanks{}
\thanks{}}

\markboth{}%
{ZHANG \MakeLowercase{\textit{et al.}}: Semi-Supervised 360 Layout Estimation with Panoramic Collaborative Perturbations}


\maketitle

\begin{abstract}
The performance of existing supervised layout estimation methods heavily relies on the quality of data annotations.
However, obtaining large-scale and high-quality datasets remains a laborious and time-consuming challenge. To solve this problem, semi-supervised approaches are introduced to relieve the demand for expensive data annotations by encouraging the consistent results of unlabeled data with different perturbations. However, existing solutions merely employ vanilla perturbations, ignoring the characteristics of panoramic layout estimation.
In contrast, we propose a novel semi-supervised method named \textbf{SemiLayout360}, which incorporates the priors of the panoramic layout and distortion through collaborative perturbations.
Specifically, we leverage the panoramic layout prior to enhance the model's focus on potential layout boundaries. Meanwhile, we introduce the panoramic distortion prior to strengthen distortion awareness. 
Furthermore, to prevent intense perturbations from hindering model convergence and ensure the effectiveness of prior-based perturbations, we divide and reorganize them as panoramic collaborative perturbations.
Our experimental results on three mainstream benchmarks demonstrate that the proposed method offers significant advantages over existing state-of-the-art (SoTA) solutions.

\end{abstract}

\begin{IEEEkeywords}
semi-supervised, collaborative perturbations, panoramic layout estimation
\end{IEEEkeywords}

\section{Introduction}
The monocular panoramic layout estimation task aims to reconstruct the 3D room layout from a single panoramic image. Room layout is one of the fundamental representations of indoor scenes, which can be parameterized by points and lines that describe the room corners and wall boundaries. This high-quality layout representation plays an important role in various applications, such as floor plan estimation \cite{r1}, scene understanding \cite{r2}, and robot localization \cite{r3,r4}. 

\begin{figure}[ht]
\centering
\includegraphics[scale=0.41]{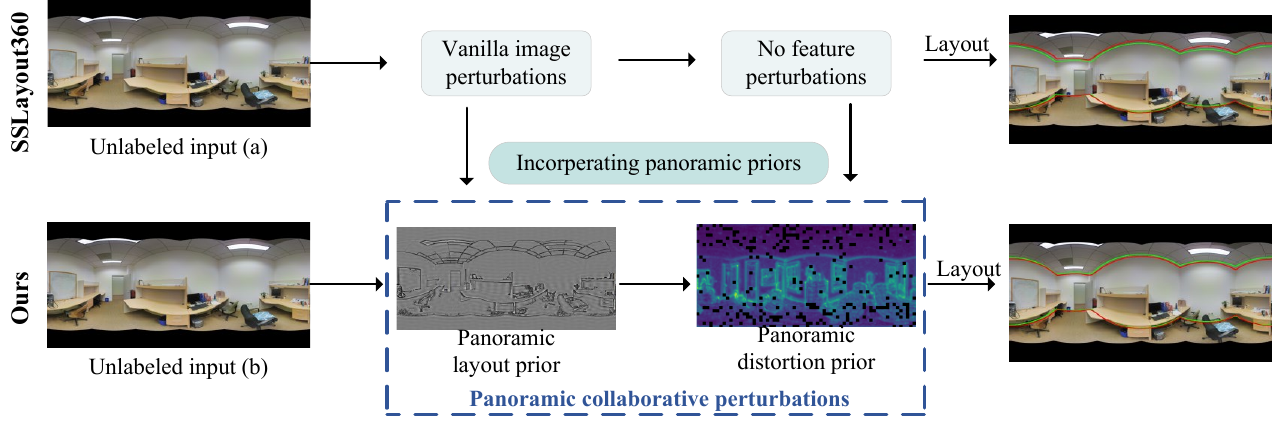}
\caption{{Brief comparisons between the previous method and our method: (a) SSLayout360 \cite{r14}, based on consistency regularization, applies vanilla perturbations (e.g., stretch, flip, rotate, gamma correction) at the image level. (b) We integrate panoramic layout and distortion priors into the perturbations and refine them into panoramic collaborative perturbations, which enables prior-based perturbations to complement each other, significantly improving the performance of semi-supervised panoramic layout estimation.}}
\label{fig0}
\end{figure}

Existing panoramic layout estimation methods largely rely on supervised learning. Some methods, such as \cite{r5,r6,r7,r8,r9}, estimate the layout from 1D sequences by compressing the extracted 2D feature maps along the height dimension to obtain the 1D sequence, where each element shares the same degree of distortion. 
To overcome the semantic confusion between different planes, DOPNet \cite{r10} decouples this 1D representation by pre-dividing orthogonal views.
Additionally, other researchers have focused on adopting different projection formats to improve performance, such as bird’s-eye view projections of rooms \cite{r11} and cube map projections \cite{r12}. These projection-based approaches effectively relieve the negative impact of image distortion.
However, the widely used panoramic layout estimation dataset, MatterportLayout \cite{r11}, proposed by Zou et al., still requires extensive manual data annotation, which demands high quality and is also time-consuming and labor-intensive. Moreover, due to the sparseness and topology of the layout estimate, completely unsupervised layout estimation is impractical in practice \cite{r13}. 

Therefore, researches on semi-supervised 360 layout estimation (SS360LE) \cite{r14,r15} have become increasingly popular. In these studies, models rely on few labeled images and numerous unlabeled images, which are from the same data distribution (such as indoor scenes). The key challenge is effectively leveraging extensive unlabeled images to approach or achieve  the performance of fully supervised methods.
The current SS360LE methods, such as SSLayout360 \cite{r14}, adopt the Mean-Teacher \cite{r47} framework based on consistency regularization. In this framework, the student model learns from labeled data in a supervised manner, while the teacher model generates soft unsupervised targets by applying random perturbations to unlabeled data. The consistency constraints between the student and teacher predictions ensure that the student model effectively learns meaningful representations from the unlabeled data.
However, As presented in the Fig. \ref{fig0} 
(a), it entirely depends on vanilla perturbations, overlooking the inherent priors of panoramic layout estimation, such as edge-concentrated layout boundaries and the non-uniform distribution of panoramic distortion.

Notably, panoramic depth estimation studies have leveraged spherical geometric priors to enhance 360 vision \cite{r71,r72}. However, SS360LE research has yet to fully exploit structural priors, such as high-frequency layout boundaries and non-uniform distortion distributions.
Therefore, we dig into task priors and incorporate them into perturbations, reformulating a new SS360LE solution named \textbf{SemiLayout360}.
Specifically, as shown in  Fig. \ref{fig0} (b), we leverage the panoramic layout prior by applying high-frequency boundary information enhancement to the input panoramic images. Afterward, we exploit panoramic distortion prior by explicitly integrating distortion-aware spatial distribution.
Guided by the priors, the two perturbations respectively highlight the structural cues of the layout and the perception of distortions in panoramic images.
However, these perturbations are overly intense, hindering model convergence. To ensure the two prior-based perturbations work effectively, we reorganize them as panoramic collaborative perturbations, balancing the two perturbations at the image and feature levels in parallel. By dynamically adjusting their magnitudes during training, the two perturbations enhance each other’s robustness without disrupting convergence.

To validate the effectiveness of our method, we conduct experiments on three widely used panoramic layout datasets: PanoContext \cite{r16}, Stanford2D3D \cite{r17}, and MatterportLayout \cite{r11}. The experimental results show that the proposed method outperforms existing state-of-the-art(SoTA) methods in both qualitative and quantitative evaluations.
The main contributions of our work are summarized as follows:

\begin{itemize}
\item We propose an SS360LE model that integrates priors into perturbations. The first prior enhances panoramic layouts through spatial-frequency augmentation to sharpen structural boundaries, while the second prior considers panoramic distortion via
distortion-aware spatial mask
\item Applying intense prior-based perturbations simultaneously can hinder model convergence. To address this, we reorganize them into panoramic collaborative perturbations, which boost each other's perturbation effectiveness without affecting convergence.
\item On multiple popular benchmarks, our method achieves better performance compared to existing state-of-the-art methods in the SS360LE task. 

\end{itemize}

\section{Related Work}
\subsection{Indoor layout estimation}
For perspective images, Zhang et al. \cite{r67} train a deconvolution network to refine edge maps for accurate room layout estimation, utilizing adaptive sampling to enhance predictions. They later propose an end-to-end framework \cite{r68} that directly predicts room layouts using transfer learning and GAN-based domain adaptation. Yan et al. \cite{r69} introduce a fully automatic method that extracts room structure lines and optimizes layout topology, enabling accurate 3D room reconstruction.

For panoramic images, many researchers have used convolutional neural networks (CNNs) to extract key features and improve the accuracy of layout estimation. For example, Zou et al. \cite{r18} propose LayoutNet, which directly predicts probability maps of corners and boundaries from the entire panorama and generates the final prediction by optimizing layout parameters. Later, they improved this method and introduced LayoutNet v2 \cite{r19}, which showed significant performance improvements over the original versions \cite{r11}. Yang et al. \cite{r20} propose DuLa-Net, which uses both equirectangular views and ceiling views to predict 2D-floor plan semantic masks. Meanwhile, Fernandez et al. \cite{r21} propose using equirectangular convolutions to generate probability maps of corners and edges. Sun et al. propose HorizonNet \cite{r5} and HoHoNet \cite{r6}, which simplify the room layout estimation process through a 1D representation. Additionally, they use Bi-LSTM and multi-head self-attention to capture long-range dependencies and refine the 1D sequences. Rao et al. \cite{r22} build their network based on HorizonNet \cite{r5}. They replace standard convolutions with spherical convolutions to reduce distortion and adopt Bi-GRU to reduce computational complexity. Wang et al. \cite{r9} integrate geometric cues of the entire layout and propose LED2-Net, re-formulating room layout estimation as predicting the depth of the walls in the horizontal direction. Pintore et al. \cite{r8} extend their work beyond Manhattan scenes and introduce AtlantaNet, which predicts room layouts by combining two projections of the floor and ceiling planes. These methods \cite{r22,r5,r9,r8}, which recover layouts from 1D sequences, have achieved impressive performance. However, compressing information into 1D sequences can obscure the semantics of different planes, leading to poorer performance and less interpretable results. In contrast, DOPNet \cite{r10} captures clear geometric cues for indoor layout estimation by pre-segmenting orthogonal planes. With the advancement of self-attention mechanisms, many transformer-based methods have been proposed to model long-range dependencies \cite{r23,r24,r25}. For instance, Jiang et al. \cite{r7} use horizon depth and room height to represent room layouts and introduce a Transformer to enhance the network's ability to learn geometric relationships. Zhang et al. \cite{r66} introduce the comprehensive depth map to planar depth conversion, which improves the problem of occlusions and position dependency.
\cite{r70} estimate acoustic 3D room structures using 360 stereo images based on cuboid modeling and semantic segmentation.

\subsection{Semi-supervised learning}
The core objective of semi-supervised learning is to fully explore and utilize the information in unlabeled data when the labeled data is limited. To achieve this goal, there are three main strategies:

The first strategy is the "pretraining-finetuning paradigm." In this approach, the neural network model is pre-trained on large-scale unlabeled data using unsupervised \cite{r26,r27}or self-supervised \cite{r28,r29} methods to learn more general feature representations. Subsequently, the model is fine-tuned using the limited labeled data to improve its performance on specific tasks.

The second strategy is "entropy minimization" \cite{r31,r32,r33,r34,r35,r36}, which is an extension of self-training \cite{r32}. This method assigns pseudo-labels to unlabeled data, reducing the model's prediction uncertainty on unlabeled data and performing end-to-end joint training using both pseudo-labels and ground truth labels. Such semi-supervised learning algorithms introduce an additional loss term into the supervised learning objective function to achieve regularization. In recent years, some self-supervised regularization methods \cite{r37,r38} have made significant progress. These methods incorporate pre-training tasks, such as image rotation recognition, as auxiliary self-supervised losses, and train them jointly with supervised image classification tasks, effectively improving the performance of image classification.

The third strategy is consistency regularization \cite{r39,r40,r41,r42,r43,r44,r45,r46,r47,r48,r49}. It aims to ensure the model's robustness to perturbed inputs, \textit{i.e.}, the model should output consistent predictions when the input is subjected to different forms of perturbations (such as noise, perturbations, \textit{etc.}). Encouraging the model to maintain consistency under these variations can improve its generalization ability. Specifically, the teacher-student framework has been widely studied in semi-supervised learning. Rasmus \textit{et al.} \cite{r50} demonstrate the effectiveness of adding random noise to the model for regularizing the objective. Miyato \textit{et al.} \cite{r45,r51} extend this idea by using adversarial noise as an implicit teacher to enhance the robustness of the model. Laine and Aila \cite{r43} adopt an exponential moving average (EMA) approach to accumulate multiple predictions, reducing the variance of the teacher's predictions. Additionally, Tarvainen and Valpola \cite{r47} propose calculating an explicit “Mean Teacher” through the EMA of the model weights, which performs well in semi-supervised learning for image classification tasks \cite{r52}. Especially when labeled samples are limited, extensions of the teacher-student framework have demonstrated stronger performance compared to fully supervised baseline methods \cite{r39,r40}. 

In this paper, we apply consistency regularization to improve the panoramic layout estimation task based on the classic Mean-Teacher \cite{r47} semi-supervised learning framework. Specifically, we employ prior-based perturbations to both the input data and extracted features, encouraging the model to generate consistent predictions when subjected to different perturbations. The consistency regularization effectively utilizes unlabeled data and enhances the model's robustness to noise and perturbed inputs, improving the overall performance in panoramic layout estimation.

\subsection{Semi-supervised layout estimation}
SSLayout360 \cite{r14} is the earliest work to explore semi-supervised panoramic layout estimation. However, although this method has made initial progress, the perturbation strategy adopted is too general and fails to utilize the special structure cues in layout estimation and the inherent distortion characteristic in panoramic images, thereby limiting the potential for performance improvement. 

In addition, another type of method for semi-supervised 360 layout estimation (SS360Layout) using point clouds \cite{r15}, while providing more detailed spatial information, requires specialized hardware like 3D sensors for data collection. This not only increases the complexity and cost of the equipment but also introduces higher computational demands, which poses significant challenges for practical applications. Therefore, how to design an SS360Layout method customized to panoramic images without the need for additional hardware support is still a problem worthy of further study.

\section{Method}
\subsection{Preliminaries}
\label{sec3_1}
\subsubsection{Mean-Teacher framework}
The Mean-Teacher \cite{r47} framework is a widely used method in the field of semi-supervised learning. It improves the performance of the model by incorporating extensive unlabeled data and limited labeled data. The core idea is to train two models (\textit{i.e.}: teacher and student models). Both models learn collaboratively to improve accuracy. Specifically, the parameters of the teacher network are the exponential moving average (EMA) of the student network's parameters during training (EMA will be explained in detail in Section 3.2). At each step of training, the student network learns from labeled data and unlabeled data, where the input is a set of labeled input-target pairs $\left ( x_{l},y_{l} \right ) \in D_{L}$ and a set of unlabeled examples $x_{u} \in D_{U}$, with $D_{L}$ and $D_{U}$ usually sampled from the related data distribution. 
The teacher network is updated gradually through weight smoothing to generate more stable predictions.
\subsubsection{DOPNet}
In contrast to methods that directly regress boundaries \cite{r5} or perform point classification \cite{r18,r8}, DOPNet \cite{r10} follows models like LED2Net \cite{r9} and LGTNet \cite{r7}, emphasizing 3D cues. It is a state-of-the-art(SoTA) model in layout estimation based on depth representations, demonstrating superior performance in the field.
In addition, the representation based on horizontal depth is essentially a form of depth estimation. Depth cues (such as contrast, boundary structures, shadow transitions, etc.) can provide useful prior information for applying perturbations.
Furthermore, a detailed comparison of different layout prediction methods in semi-supervised learning is not covered in this paper but could be an interesting topic for future research.

DOPNet takes a 512×1024 panoramic image as input, using ResNet to extract features at four scales. Multi-scale feature fusion reduces distortion and improves layout accuracy. The soft-flipping strategy leverages room symmetry to capture global features. Finally, the model generates an accurate estimation of horizon-depth and room height, thus achieving precise room layout prediction.



\begin{figure*}[ht]
\centering
\includegraphics[scale=0.48]{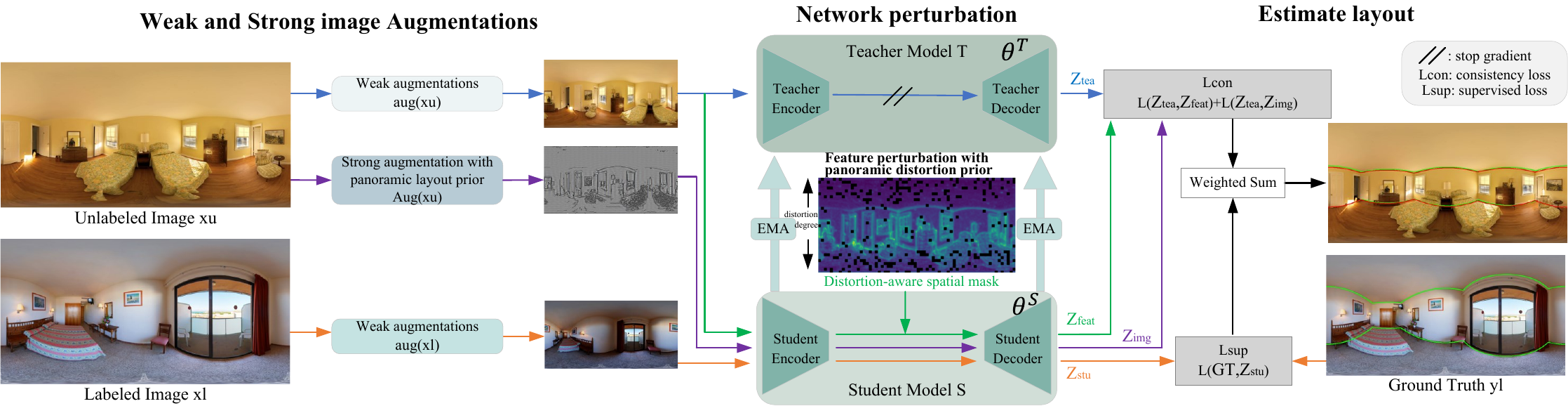}
\caption{ {Overview of the framework of SemiLayout360. In the standard teacher-student framework, SemiLayout360 trains the student model S(parameterized by $\theta ^{S}$ ) on both labeled data $\left ( x_{l},y_{l} \right )$ and unlabeled data $x_{u}$, by minimizing the corresponding supervised loss $L_{sup}$ and unsupervised consistency loss $L_{con}$. The teacher model (parameterized by $\theta ^{T}$  ) is updated via the exponential moving average (EMA) of the student model's parameters and generates pseudo-labels $Z_{tea}$ for the unlabeled data. The core of SemiLayout360 is to apply multiple perturbations on the unlabeled samples, including image, feature, and network perturbations.}}
\label{fig1}
\end{figure*}

\subsection{Architecture Overview}
In Fig. \ref{fig1}, we show our complete framework motivated
by the panoramic priors in a semi-supervised setting. 
Overall, SemiLayout360 integrates image and feature perturbations to improve robustness and accuracy in estimating layout with the student-teacher framework.
We design DOPNet to function in two capacities: as both student and teacher models. Given a batch of labeled samples, augmented as $aug(x_{l})$, with their corresponding ground truth labels $y_{l} \in R^{3\times 1\times 1024}$, along with a batch of augmented unlabeled samples $aug(x_{u})$, $Aug(x_{u})$, we perform a forward pass of DOPNet three times: 

(1) \textbf{Student Model (S):} 
On the labeled sample batch, the model is trained as the student to generate real-valued prediction vectors $Z_{stu} \in R^{3\times 1\times 1024}$. In this step, the student model learns from the labeled data to estimate layout information.    

(2) \textbf{Image Perturbation and Feature Perturbation:}
On the unlabeled sample batch, we apply both image perturbation and feature perturbation. From this, we obtain $Z_{img} \in R^{3\times 1\times 1024}$   and $Z_{feat} \in R^{3\times 1\times 1024}$. These perturbations help the model learn more robust and generalized features.

(3) \textbf{Teacher Model (T):} 
The model is passed through as the teacher, where it outputs pseudo-labels $Z_{tea}$, using the same unlabeled batch. These pseudo-labels are then used to guide the student during training further, facilitating the semi-supervised learning process.


\subsection{Multiple Perturbations}
\subsubsection{Input image perturbation}
We take labeled image-label pairs and unlabeled images as input.
As shown in the image perturbations in Fig. \ref{fig1}, We divide image perturbations into vanilla weak augmentation and prior-based strong augmentation. For weak augmentation, we follow common panoramic image enhancement strategies \cite{r5,r53,r18,r10}.  Specifically, the operations include left-right flipping with a probability of 0.5, random panoramic stretching in the range $\left ( k_{x},k_{z}  \right ) \in \left [ 0.5,2 \right ]$, and panoramic horizontal rotation $r\in \left ( 0^{\circ },360^{\circ } \right )$. 

For strong augmentation, we enhance the model's ability to perceive high-frequency boundaries by integrating the panoramic layout prior. Specifically, we use histogram equalization to strengthen the brightness and contrast of the images.
Afterwards, to further enhance the geometric details and boundary structure, we use the Fourier transform to apply a high-pass filter in the frequency domain, suppressing low-frequency components and emphasizing high-frequency information such as edges and contours. This approach effectively strengthens the structural boundaries in indoor panoramic scenes, allowing the model to capture important features better when handling complex panoramic layouts.

In particular, these strong augmentations are only applied to the images processed by the student model. In this way, the student model is able to cope with more challenging input images during training, improving its understanding and adaptability to complex scenes and panoramic distortions.

\subsubsection{Feature perturbation}
For weak augmented images, we apply a specially designed feature perturbation technique after extracting features through the encoder. Our approach introduces a spatial mask with a structured probability distribution, where the mask probability is higher in the edge regions of the feature map and lower in the center. This is based on the assumption that distortion is more severe in the edge regions of panoramic images, therefore larger perturbation in training can improve the robustness of the model.

First, we generate a spatial mask that distinguishes the center and edge regions of the feature map. To achieve this, we define a probability gradient that increases from the center of the image to the edges. Specifically, the mask probability of the center of the feature map is defined as $P_{center}$, while the mask probability of the edges is defined as $P_{edge}$. The probability transformation between the center and edge regions follows a distance-based quadratic function relationship:
\begin{equation}
P\left ( y \right ) =P_{center} +\left ( P_{edge} -P_{center}\right ) \times y^{2} 
\label{equ1}
\end{equation}
where $y$ is a normalized coordinate representing the vertical position of the image, ranging from -1 to 1. $P_{center}$ and $P_{edge}$ are set as 0.8 and 0.2 according to experiments (in Table \ref{table4}).

To further introduce diversity in the feature space, we selectively mask some channels. Through a channel-level mask probability $P_{channel}$, 20\% of the channels are randomly selected for masking. For the selected channels, the previously defined spatial mask is applied, while the unselected channels remain unchanged. When using the mask for feature perturbation, some features are randomly set to zero, resulting in a decrease in the total amount of activation values in the feature map. This, consequently, impairs the network's capacity to transmit information.
To alleviate this issue, we design a scaling strategy based on the proportion of retained features, introducing a dynamically adjustable scaling factor to compensate for the reduction in activation values caused by the discarded features. Let $P_{f}$ denotes the proportion of retained features after applying the mask, representing the proportion of non-zero elements during masking. The scaling factor $S$ is defined as $S=\frac{1}{P_{f}}$. When $P_{f}$ is low, it means that more features have been masked, so $S$ increases accordingly to amplify the activation values of the remaining features, ensuring that the overall output magnitude of the feature map is kept at the same level as before the perturbation.

\subsubsection{Network perturbation}
The network perturbation primarily originates from the teacher and student models within the Mean-Teacher framework. This framework introduces perturbations between the teacher and the student model and promotes the learning of the model through consistency constraints. Specifically, it ensures that the student model’s predictions remain consistent with the outputs of the teacher model. In practice, the teacher model generates the pseudo label $Z_{tea}$, and the student model produces $Z_{feat}$ and $Z_{img}$ according to different perturbations.

During the training process, the teacher model's parameters $\theta ^{T}$   are updated by the exponential moving average (EMA) of the student model's parameters $\theta ^{S}$, with the update occurring at each training iteration:
\begin{equation}
\theta _{i}^{T} =\alpha \theta _{i-1}^{T}+(1-\alpha )\theta _{i}^{S}
\label{equ3}
\end{equation}
where $\alpha \in \left [ 0,1 \right ]$ is the decay hyper-parameter. The goal for setting $\theta ^{T} =EMA\left ( \theta ^{S}  \right )$  is to obtain a good teacher model to provide stable unsupervised targets for the student to mimic, which is the main outcome of the Mean Teacher framework. The usual practice is not to back-propagate gradients through the teacher model and to keep its predictions unchanged at each training step. In our experiments, we set $\alpha$ to be consistent with SSLayout360, choosing $\alpha$ = 0.999.
\subsubsection{Panoramic collaborative perturbations}
We refine the image and feature perturbations based on the panoramic priors into panoramic collaborative perturbations, aiming to enhance the robustness and generalization ability of the student model. Specifically, for weakly augmented images, we apply feature perturbation after the encoder to further increase the diversity of the learned features. In contrast, for strongly augmented images, we do not use feature perturbation. Experimental results show that continuing to apply feature perturbation on strongly augmented images will reduce model performance (in Table \ref{table3}). This is because strong augmentation techniques, such as histogram equalization and Fourier transform, have already made the edges and contours of the images more pronounced, which is crucial for accurate panoramic layout prediction. Excessive feature perturbation will lead to the loss of key layout information, causing the model to deviate from core information.

\subsection{loss function}
In this work, we design the loss function consisting of two main parts: an unsupervised consistency loss $L_{con}$ based on unlabeled data and a supervised loss $L_{sup}$ based on labeled data. The total loss function is a weighted combination of the two parts.

The unsupervised consistency loss $L_{con}$ comes from the prediction consistency between the teacher and the student model on the unlabeled data. Specifically, the teacher model generates the pseudo label $Z_{tea}$ for the unlabeled data to guide the learning of the student model. The student model produces two outputs for the unlabeled data: a feature-perturbed $Z_{feat}$ and an image-perturbed output $Z_{img}$. The consistency loss is calculated by comparing the difference between the student model's predictions $Z_{feat}$ and $Z_{img}$ and the pseudo label $Z_{tea}$ of the teacher model. This loss encourages the student model to keep similar predictions under different perturbations, thereby improving the model's robustness to input variations.

\begin{equation}
L_{con} =L\left ( Z_{tea},Z_{feat}  \right ) +L\left ( Z_{tea},Z_{img}  \right ) 
\label{equ4}
\end{equation}
The supervised loss $L_{sup}$ is calculated using the labeled data. The student model directly generates the predicted output $Z_{stu}$, and the $L_{sup}$ is computed based on the difference between the predicted output and the ground truth (GT).
\begin{equation}
L_{sup} = L\left ( Z_{stu},GT  \right ) 
\label{equ5}
\end{equation}
The total loss function is a weighted addition of supervised and unsupervised consistency loss. The unsupervised consistency loss $L_{con}$ serves as a regularization term that enhances the learning ability of the student model by introducing the information of unlabeled data. To control the influence of the consistency loss, we introduce a weight factor $\lambda$.
Additionally, We follow DOPNet \cite{r10} and LGTNet \cite{r7} in $L_{sup}$ and $L_{con}$, and both use the following loss composition:
\begin{equation}
Loss=\alpha L_{d}+\mu  L_{h}+\nu \left ( L_{n}+L_{g} \right ) 
\label{equ6}
\end{equation}
where $L_{d}$ and $L_{h}$ represent the horizon-depth and room height losses, and we use L1 loss, both calculated using the L1 loss function.$L_{n}$ denotes the normal loss, and $L_{g}$ represents the gradient loss.
Based on empirical results, We set $\alpha$ to 0.9, $\mu$ to 0.1, and $\nu$ to [1.0, 1.0].
\begin{equation}
L_{total} = L_{sup} +\lambda L_{con}
\label{equ7}
\end{equation}
In the early stages of model training, especially when there are few available labels, the predictions of the student and the teacher model may be inaccurate and inconsistent. To alleviate this problem, we introduce a strategy called the "ramp-up period." During the ramp-up period, the weight $\lambda$ of the consistency loss for unlabeled data gradually increases from 0 to 1. The duration of the ramp-up period is controlled by a sigmoid-shaped function (as shown in   Eq. \ref{equ7}), which gradually increases with the number of training iterations.
\begin{equation}
\lambda \left ( i \right ) =e^{-5\left ( ^{}1-\frac{i}{I}   \right ) ^{2} } 
\label{equ8}
\end{equation}
Here, $i$ represents the current training iteration, and $I$ represents the iteration number when the ramp-up period ends. In this work, we define $I$ as 30\% of the maximum number of iterations based on experiments (in Table \ref{table5}). This strategy aims to ensure that the model mainly relies on labeled data for learning in the early stage of training. After the ramp-up period ends, the teacher model can provide the student model with more reliable and stable unsupervised signals, further improving the student's learning performance.

\begin{figure*}[ht]
\centering
\includegraphics[scale=0.6]{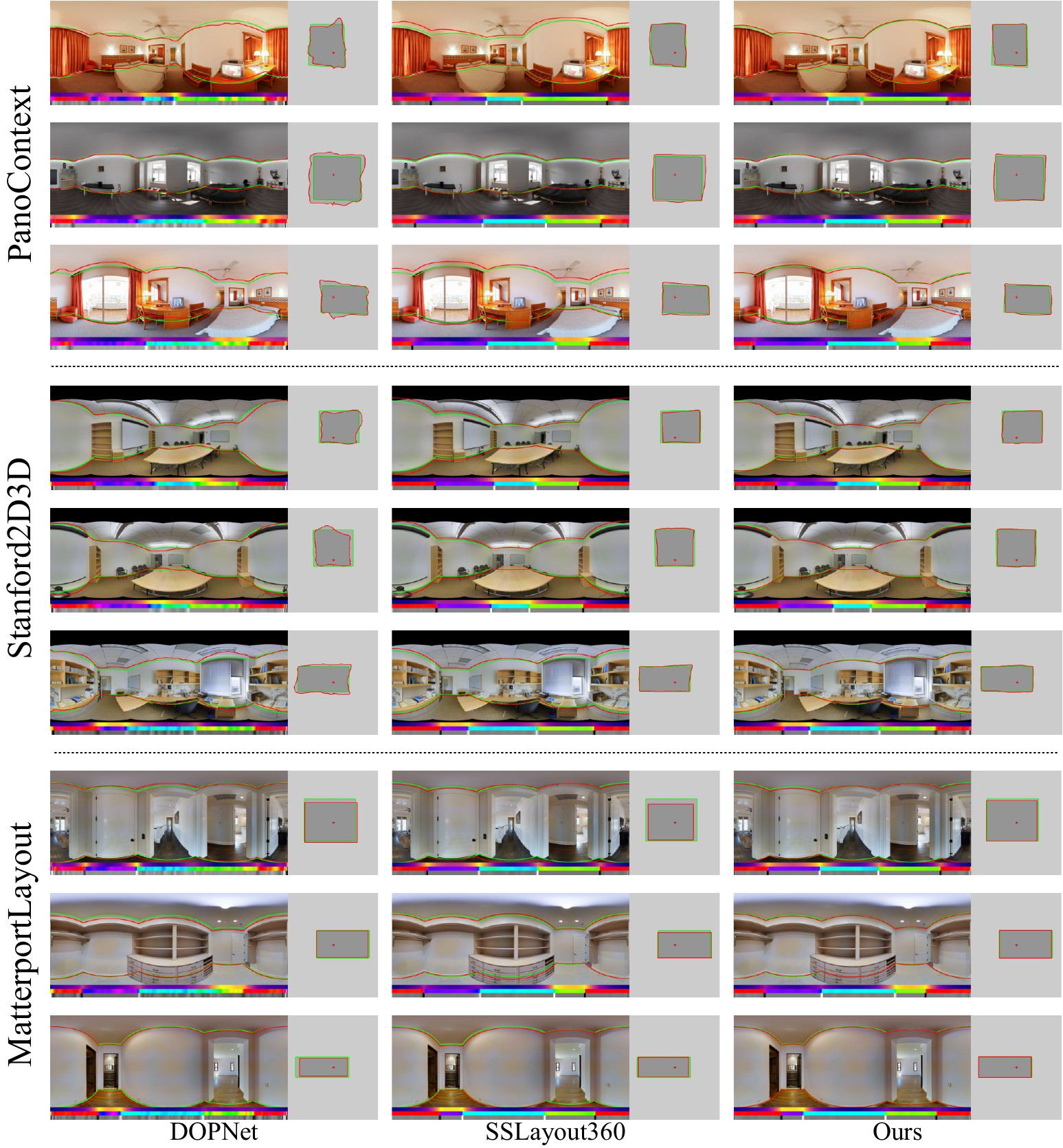}
\caption{ {Qualitative results on the PanoContext dataset (top), Stanford2D3D dataset (middle), and MatterportLayout dataset (bottom). We compare our SemiLayout360 with the supervised DOPNet and SSLayout360. The supervised DOPNet is trained on 100 labels, while our SemiLayout360 and SSLayout360 use the same 100 labels along with unlabeled images. The boundaries of the room layout on a panorama are shown on the left and the floor plan is on the right. Ground truth is viewed in \textcolor{green}{Green lines} and the prediction in \textcolor{red}{Red}. The predicted horizon depth, normal, and gradient are visualized below each panorama. We observe that SemiLayout360 predicts layout boundary lines following
more closely to the ground truth than DOPNet and SSLayout360,which demonstrates the effectiveness of applying customized image and feature perturbation strategies. }}
\label{fig2}
\end{figure*}

\section{Experiments}
\subsection{Datasets and Implementation Details}
\textbf{Datasets:}
Our SemiLayout360 is trained and evaluated on three benchmark datasets: Stanford2D3D \cite{r17}, PanoContext \cite{r16}, and MatterportLayout \cite{r11}. PanoContext and Stanford2D3D are two commonly used datasets for indoor panoramic room layout estimation, containing 512 and 550 cuboid room layouts, respectively. The Stanford2D3D dataset is annotated by Zou et al. \cite{r18} and has a smaller vertical field of view (FOV) compared to other datasets. In addition, the MatterportLayout dataset is a subset of the Matterport3D \cite{r54} dataset, also annotated by Zou et al. \cite{r11},  containing 2295 non-cuboid room layouts. To ensure a fair comparison, we strictly follow the same training, validation, and test splits used in prior work \cite{r14}.

\textbf{Implementation Details:}
In both supervised and semi-supervised learning (SSL) experiments, we use the same architecture and training protocol to ensure that the performance improvements in SSL are attributed to the introduction of unlabeled data and perturbations rather than changes in model configuration. In our experimental settings, we perform all experiments using a single GTX 3090 GPU. The method is implemented using PyTorch. We choose Adam \cite{r55} as the optimizer and follow DOPNet’s training settings. The initial learning rate is $1\times 10^{-4}$, and the batch size during training is set to 4.
We save the best model for testing based on their performances on the validation set. During testing, SemiLayout360 generates two sets of model parameters: $\theta ^{S}$ and $\theta ^{T}$ = $EMA(\theta ^{S})$. We select the better result from the two test instances for reporting. Additionally, we do not perform any test-time data augmentation.

\begin{table*}[htbp]
\centering
\caption{Quantitative cuboid layout results evaluated on the PanoContext (left) and Stanford-2D3D (right) test sets. $^{\dagger }$ means that we modify the SSLayout360 model to DOPNet to ensure a fair comparison.}
\label{table1}
\begin{adjustbox}{max width=\textwidth}
\begin{tabular}{@{}lccccc|ccccc@{}}
\toprule
& \multicolumn{5}{c|}{\textbf{PanoContext \cite{r16}}} & \multicolumn{5}{c}{\textbf{Stanford2D3D \cite{r17}}} \\ \midrule
\multirow{2}{*}{\textbf{Method}} & \textbf{20 labels} & \textbf{50 labels} & \textbf{100 labels} & \textbf{200 labels} & \textbf{963 labels} & \textbf{20 labels} & \textbf{50 labels} & \textbf{100 labels} & \textbf{200 labels} & \textbf{916 labels} \\
                & \textbf{1,009 images} & \textbf{1,009 images} & \textbf{1,009 images} & \textbf{1,009 images} & \textbf{1,009 images} & \textbf{949 images} & \textbf{949 images} & \textbf{949 images} & \textbf{949 images} & \textbf{949 images} \\ \midrule
& \multicolumn{5}{c|}{\textbf{3D IoU (\%) $\uparrow$}} & \multicolumn{5}{c}{\textbf{3D IoU (\%) $\uparrow$}} \\ \midrule
DOPNet \cite{r10} & 53.58 & 54.92 & 66.25 & 73.36 & 81.37 & 59.42 & 62.68 & 73.39 & 75.91 & 79.59 \\
SSLayout360$^{\dagger }$ \cite{r14} & 62.56 & 71.62 & 73.11  & 76.42 & 82.81 & 71.45 & 74.90 & 77.62 & 79.94 & 81.50 \\
SemiLayout360 & \textbf{62.75} & \textbf{72.83} & \textbf{76.23} & \textbf{79.37} &  \textbf{84.30} & \textbf{73.66} & \textbf{75.47} & \textbf{80.29} & \textbf{80.92} & \textbf{82.76} \\ \midrule
& \multicolumn{5}{c|}{\textbf{2D IoU (\%) $\uparrow$}} & \multicolumn{5}{c}{\textbf{2D IoU (\%) $\uparrow$}} \\ \midrule
DOPNet \cite{r10} & 59.01 & 60.96 & 70.30 & 77.23 & 84.62 & 63.14 & 67.13 & 77.73 & 80.31 & 84.76 \\
SSLayout360$^{\dagger }$ \cite{r14} & 70.52 & 75.00 & 78.05  & 79.52 & 85.82 & \textbf{78.63} & 78.28 & 80.58 & 83.51 & 84.21 \\
SemiLayout360 & \textbf{70.82} & \textbf{77.11} & \textbf{79.81} & \textbf{82.99} &  \textbf{87.16} & 77.59 & \textbf{78.91} & \textbf{83.81} & \textbf{84.74} & \textbf{85.26} \\ \midrule
& \multicolumn{5}{c|}{\textbf{Corner Error (\%) $\downarrow$}} & \multicolumn{5}{c}{\textbf{Corner Error (\%) $\downarrow$}} \\ \midrule
DOPNet \cite{r10} & 2.61 & 2.54 & 1.80 & 1.31 & 0.91 & 2.28 & 2.24 & 1.31 & 1.13 & 0.96 \\
SSLayout360$^{\dagger }$ \cite{r14} & 1.75  & 1.29 & 1.28 & 1.03 & 0.87 & 1.46 & 1.14 & 0.91 & \textbf{0.84} & 0.82 \\
SemiLayout360 & \textbf{1.73} & \textbf{1.22} & \textbf{1.01} & \textbf{0.91} & \textbf{0.77} & \textbf{1.26} & \textbf{0.99} & \textbf{0.87} & 0.89 & \textbf{0.78} \\ \midrule
& \multicolumn{5}{c|}{\textbf{Pixel Error (\%) $\downarrow$}} & \multicolumn{5}{c}{\textbf{Pixel Error (\%) $\downarrow$}} \\ \midrule
DOPNet \cite{r10} & 9.64 & 9.99 & 6.19 & 4.06 & 2.77 & 8.93 & 8.29 & 4.76 & 4.35 & 3.49 \\
SSLayout360$^{\dagger }$ \cite{r14} & 6.64  & 4.41 & 4.55 & 3.28 & 2.81 & 5.40 & 4.30 & 3.45 & \textbf{2.84} & 2.89 \\
SemiLayout360 & \textbf{6.09} & \textbf{3.95} & \textbf{3.25} & \textbf{2.96} & \textbf{2.34} & \textbf{4.09} & \textbf{3.33} & \textbf{3.03} & 3.03 & \textbf{2.59} \\ \bottomrule
\end{tabular}
\end{adjustbox}
\end{table*}

\subsection{Comparison Results }
\textbf{Metrics:}
To evaluate the SSL performance fairly, we select a series of standard evaluation metrics in SSLayout360 \cite{r14}. We evaluate cuboid layouts by 3D intersection over union (3D IoU), 2D IoU, corner error (CE), and pixel error (PE). For non-cuboid layouts, we evaluate using 3D IoU, 2D IoU, root mean squared error (RMSE), and $\delta 1$. $\delta 1$ is described by Zou et al. \cite{r11}  as the percentage of pixels where the ratio between the prediction depth and ground truth depth is within a threshold of 1.25.

\textbf{Quantitative Analysis:}
In Table \ref{table1}, we present the quantitative comparison results for cuboid layout estimation on the PanoContext \cite{r16} and Stanford-2D3D \cite{r17} datasets. Our SemiLayout360 outperforms the supervised DOPNet baseline and the semi-supervised SSLayout360 in nearly all metrics. For the fully supervised setting with all labeled images, SemiLayout360 surpasses the supervised DOPNet baseline in 3D and 2D IoU metric, and also performs better on the corner error and pixel error metrics. This demonstrates the benefits of incorporating consistency regularization for layout estimation.
In Table \ref{table2}, we provide the quantitative comparison results for Non-Cuboid layout estimation on the MatterportLayout \cite{r11} dataset. Similarly, SemiLayout360 outperforms the supervised DOPNet baseline and the semi-supervised SSLayout360 in almost all metrics.

\textbf{Qualitative Analysis:}
In Fig. \ref{fig2}, we compare the qualitative test results of DOPNet \cite{r10}, SSLayout360 \cite{r14}, and our SemiLayout360, all trained on 100 labels, across the PanoContext \cite{r16}, Stanford2D3D \cite{r17}, and MatterportLayout \cite{r11} datasets under the equirectangular view. From the figure, it can be observed that our method achieves more accurate boundaries of the room layout. Additionally, the visualizations of floor plans demonstrate that our approach provides better results, benefiting from the prior-based image and feature perturbation strategies.

\begin{table}[htbp]
\centering
\caption{Quantitative non-cuboid layout results evaluated on the MatterportLayout test set. $^{\dagger }$ means that we modify the SSLayout360 model to DOPNet to ensure a fair comparison.}
\resizebox{0.5\textwidth}{!}{ 
\begin{tabular}{l|cccccc}
\toprule
\multicolumn{6}{c}{\textbf{MatterportLayout \cite{r11}}} \\ \midrule
\multirow{2}{*}{\textbf{Method}} & \textbf{50 labels} & \textbf{100 labels} & \textbf{200 labels} & \textbf{400 labels} & \textbf{1650 labels} \\ 
                & \textbf{1,837 images} & \textbf{1,837 images} & \textbf{1,837 images} & \textbf{1,837 images} & \textbf{1,837 images} \\ \midrule
\multicolumn{6}{c}{\textbf{3D IoU (\%) ↑}} \\ \midrule
DOPNet \cite{r10}          & 63.45 & 68.90 & 74.22 & 76.54 & 79.08 \\
SSLayout360$^{\dagger }$ \cite{r14}     & 72.22 & 73.80 & 78.16 & 79.71 & 80.14 \\
SemiLayout360         & \textbf{73.19} & \textbf{76.74} & \textbf{79.29} & \textbf{80.43} & \textbf{80.77} \\ \midrule
\multicolumn{6}{c}{\textbf{2D IoU (\%) ↑}} \\ \midrule
DOPNet \cite{r10}          & 68.38 & 72.45 & 77.09 & 79.58 & 81.71 \\
SSLayout360$^{\dagger }$ \cite{r14}     & 75.65 & 77.32 & 80.56 & 82.29 & 82.69 \\
SemiLayout360         & \textbf{77.15} & \textbf{79.49} & \textbf{81.59} & \textbf{82.71} & \textbf{83.24} \\ \midrule
\multicolumn{6}{c}{\textbf{$\delta 1$ ↑}} \\ \midrule
DOPNet \cite{r10}          & 0.7245 & 0.8175 & 0.8961 & 0.9197 & 0.9432\\
SSLayout360$^{\dagger }$ \cite{r14}     & \textbf{0.8757} & 0.8944 & 0.9374 & 0.9476 & \textbf{0.9501} \\
SemiLayout360         & 0.8731 & \textbf{0.9151} & \textbf{0.9460} & \textbf{0.9523} & 0.9481\\ \midrule
\multicolumn{6}{c}{\textbf{RMSE  ↓}} \\ \midrule
DOPNet \cite{r10}          & 0.4068 & 0.3340 & 0.2695 & 0.2458 & 0.2217 \\
SSLayout360$^{\dagger }$ \cite{r14}     & 0.3129 & 0.2847 & 0.2337& 0.2143 & 0.2121 \\
SemiLayout360         & \textbf{0.2978} & \textbf{0.2471} & \textbf{0.2185} & \textbf{0.2056} & \textbf{0.2026} \\
\bottomrule
\end{tabular}
}
\label{table2}
\end{table}

\begin{figure*}[ht]
\centering
\includegraphics[scale=0.6]{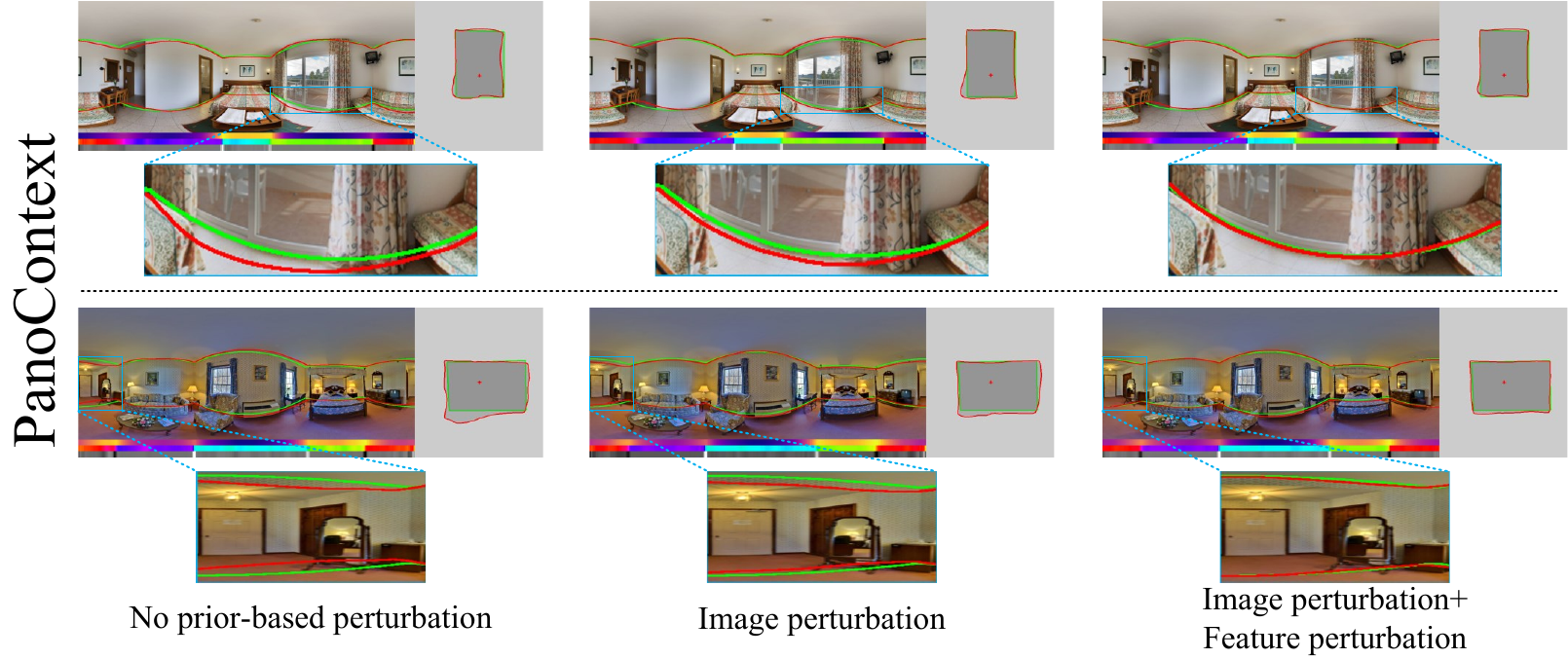}
\caption{ {Qualitative comparisons about individual perturbation on the PanoContext dataset. As we add image and feature perturbations from left to right collaboratively, the boundaries of the room layout become more accurate. Ground truth is viewed in \textcolor{green}{Green lines} and the prediction in \textcolor{red}{Red} }}
\label{fig3}
\end{figure*}

\begin{table*}[htbp]
\centering
\caption{Ablation studies of individual components.
We begin without applying any perturbation and gradually add image and feature perturbations. When both image and feature perturbations are applied together, the panoramic collaborative perturbation is used. We conduct a series of ablation studies on the PanoContext dataset (trained with 100 labels), where the results in bold indicate the best performance.) }
\label{table3}
\begin{adjustbox}{max width=\textwidth}
{
\begin{tabular}{ c|c|c|c|c|c|c|c|c}
\hline
  Data & ID & Image perturbation & Feature perturbation & Collaborative perturbation & 3D IoU (\%) $\uparrow$ & 2D IoU (\%) $\uparrow$ & Corner Error (\%) $\downarrow$ & Pixel Error (\%) $\downarrow$   \\
\hline
\multirow{4}*{PanoContext \cite{r16}}
&a) &\XSolidBrush &\XSolidBrush &--- &72.62 &77.05 &1.23  &4.41    \\
&b) &\Checkmark &\XSolidBrush &--- &73.88 &77.69 &1.13  &3.58   \\
&c) &\Checkmark &\Checkmark &\XSolidBrush &69.34  &73.65 &1.49 &5.16   \\
&d) &\Checkmark &\Checkmark &\Checkmark &\textbf{76.23}  &\textbf{79.81} &\textbf{1.01} &\textbf{3.25}   \\
\hline
\end{tabular}
}
\end{adjustbox}
\end{table*}

\subsection{Ablation study}
We conduct a thorough validation of the key components of our SemiLayout360 under the same experimental conditions. As shown in Table \ref{table3}, we initially do not apply any prior-based perturbations and instead enforce consistency constraints by applying different data augmentations to the inputs of the student and teacher models. Subsequently, we sequentially introduce image perturbation based on panoramic layout prior and feature perturbation based on panoramic distortion prior, then refine these into panoramic collaborative perturbations using the PanoContext dataset (trained with 100 labels) to evaluate the effectiveness of our SemiLayout360.

\subsubsection{Effectiveness of image perturbation}
From Table \ref{table3}, we can observe that on the PanoContext dataset, all metrics improve after applying image perturbations. Furthermore, as shown in Fig. \ref{fig3}, the estimation of room layouts becomes more accurate with image perturbations. These results demonstrate that by incorporating image perturbations, such as histogram equalization and Fourier transform, the contrast and details of the images are enhanced, strengthening the boundary structure, and capturing the important features in indoor panoramic scenes.

\subsubsection{Effectiveness of feature perturbation}
The introduction of a spatial mask with a structured probability distribution takes into account the distortion distribution prior in panoramic images. The results in Table \ref{table3} show that the main metrics improve after applying feature perturbation. As shown in Fig. \ref{fig3}, the predicted layout becomes more accurate with the addition of feature perturbation.

\subsubsection{Effectiveness of panoramic collaborative perturbations}Pnoramic collaborative perturbations are introduced to avoid the negative impact of intense perturbations on model convergence while ensuring the effectiveness of prior-based perturbations.
Comparisons in Table \ref{table3} reveal a 7.59\% improvement (3DIoU) and 6.99\% (2DIoU) on the PanoContext dataset.

\subsubsection{Mask ratio}
We set different initial mask ratios for $P_{edge}$ and $P_{center}$ based on the distortion characteristic of panoramic images and conduct experimental comparisons, as shown in Table \ref{table4}. The best performance was achieved when $P_{center}$ is 0.2 and $P_{edge}$ is 0.8.

\begin{table}[htbp]
\centering
\caption{Performance comparison with different initial values of \texttt{$P_{edge}$} and \texttt{$P_{center}$}. CE denotes Corner Error and PE represents Pixel Error}
\resizebox{0.5\textwidth}{!}{
\begin{tabular}{|c|c|c|c|c|c|}
\hline
\multirow{2}{*}{\textbf{$P_{edge}$}} & \multirow{2}{*}{\textbf{$P_{center}$}} & \multicolumn{4}{c|}{\textbf{Performance Metrics}} \\ \cline{3-6} 
& & \textbf{3D IoU (\%) $\uparrow$} & \textbf{2D IoU (\%) $\uparrow$} & \textbf{CE (\%) $\downarrow$} & \textbf{PE (\%) $\downarrow$} \\ \hline
\multirow{3}{*}{0.7} & 0.1 & 74.59 & 78.78 & 1.10 & 3.59 \\ \cline{2-6} 
 & 0.2 & 74.34 & 78.26 & 1.06 & 3.43 \\ \cline{2-6} 
 & 0.3 & 74.59 & 78.74 & 1.02 & 3.52 \\ \hline
\multirow{3}{*}{0.8} & 0.1 & 75.15 & 79.08 & 1.09 & 3.54 \\ \cline{2-6} 
 & 0.2 & \textbf{76.23} & \textbf{79.81} & \textbf{1.01} & \textbf{3.25} \\ \cline{2-6} 
 & 0.3 & 75.64 & 79.63 & 1.07 & 3.64 \\ \hline
\multirow{3}{*}{0.9} & 0.1 & 74.28 & 78.68 & 1.12 & 3.79 \\ \cline{2-6} 
 & 0.2 & 75.15 & 77.24 & 1.49 & 4.83 \\ \cline{2-6} 
 & 0.3 & 74.58 & 78.23 & 1.13 & 3.71 \\ \hline
\end{tabular}
}
\label{table4}
\end{table}

\subsubsection{Ramp-up period}We set different ramp-up period termination ratios, as shown in Table \ref{table5}. The best performance is achieved when termination occurs at 30\% of the total iterations.

\begin{table}[htbp]
\centering
\caption{Performance comparison with different termination ratios during the ramp-up period. CE denotes Corner Error, and PE represents Pixel Error}
\resizebox{0.5 \textwidth}{!}{
\begin{tabular}{|c|c|c|c|c|}
\hline
\textbf{Termination Ratio} & \textbf{3D IoU (\%) $\uparrow$} & \textbf{2D IoU (\%) $\uparrow$} & \textbf{CE (\%) $\downarrow$} & \textbf{PE (\%) $\downarrow$} \\ \hline
\textbf{10\%} & 70.61 & 74.93 & 1.37 & 4.55 \\ \hline
\textbf{30\%} & \textbf{76.23} & \textbf{79.81} & 1.01 & \textbf{3.25} \\ \hline
\textbf{50\%} & 74.49 & 78.29 & 1.01 & 3.26 \\ \hline
\textbf{70\%} & 73.67 & 76.93 & 1.03 & 3.44 \\ \hline
\end{tabular}
}
\label{table5}
\end{table}

\vspace{-10pt}
\section{Conclusion}
In this paper, we propose a novel semi-supervised method
for monocular panoramic layout estimation, SemiLayout360, which integrates panoramic priors into perturbations. Considering the characteristics of the layout estimation task, we first leverage the panoramic layout prior and apply histogram equalization to strengthen the brightness and contrast of the scene. We then use the Fourier transform to highlight the boundaries. 
Due to the inherent distortion distribution of panoramic images, we design a distortion-aware spatial mask using the panoramic distortion prior to improve the robustness in the polar regions, where distortion is more significant. Additionally, we refine prior-based perturbations into panoramic collaborative priors, which can enhance each other's perturbation effectiveness without hindering model convergence. Experiments on three benchmarks
demonstrate that SemiLayout360 significantly outperforms SoTA methods.

\bibliographystyle{IEEEtran}
\bibliography{ref}
\end{document}